\documentclass{Interspeech2024}
\usepackage{amsmath}
\usepackage{multirow}

\interspeechcameraready

\newcommand{\datasethobbit}{B2B-1}
\newcommand{\datasetkorra}{B2B-2}
\newcommand{\datasetthk}{B2B-3}

\title{Multimodal Large Language Models with Fusion Low Rank Adaptation for Device Directed Speech Detection}

\name{Shruti Palaskar, Oggi Rudovic, Sameer Dharur, Florian Pesce, Gautam Krishna, Aswin Sivaraman, Jack Berkowitz, Ahmed Hussen Abdelaziz, Saurabh Adya, Ahmed Tewfik}

\address{Apple}
\email{spalaskar@apple.com}


\begin{document}

\maketitle

\begin{abstract}
Although Large Language Models (LLMs) have shown promise for human-like conversations, they are primarily pre-trained on text data. Incorporating audio or video improves performance, but collecting large-scale multimodal data and pre-training multimodal LLMs is challenging. To this end, we propose a Fusion Low Rank Adaptation (FLoRA) technique that efficiently adapts a pre-trained unimodal LLM to consume new, previously unseen modalities via low rank adaptation. For device-directed speech detection, using FLoRA, the multimodal LLM achieves 22\% relative reduction in equal error rate (EER) over the text-only approach and attains performance parity with its full fine-tuning (FFT) counterpart while needing to tune only a fraction of its parameters. Furthermore, with the newly introduced adapter dropout, FLoRA is robust to missing data, improving over FFT by 20\% lower EER and 56\% lower false accept rate. The proposed approach scales well for model sizes from 16M to 3B parameters. 
\end{abstract}

\section{Introduction}
\label{sec:introduction}

Recent breakthroughs in speech and language technologies, multimodal human-machine interaction, and the introduction of Large Language Models (LLMs) have facilitated the adoption of Voice Assistants (VAs) as a pervasive interface for smart home control and information retrieval tasks. While VAs exhibit optimal performance in controlled acoustic environments, the presence of background noise and ambiguous boundaries between intended commands and ambient noise pose significant challenges for their accuracy and reliability in real-world applications. To mitigate these limitations and to optimize user experience, the development of a Device Directed Speech Detection (DDSD) system \cite{mallidi2018device} that is capable of effectively distinguishing device-directed queries from side conversations and ambient noise is crucial. Such a system can enhance the VA performance in real-world scenarios, ultimately paving the way for seamless and reliable human-machine interaction in diverse acoustic environments.

While LLMs show remarkable zero-shot learning capabilities, enabling task adaptation through prompting without further training \cite{agrawal-etal-2022-large,NEURIPS2020_1457c0d6,Wang2023CanCW,Kojima2022LargeLM,dighe2023leveraging}, their deployment for a task, such as DDSD, faces some challenges. Firstly, real-world applications with Automatic Speech Recognition (ASR) can be prone to misinterpretations due to acoustic artifacts or background noise. Secondly, even in controlled environments, relying solely on transcribed text without additional context from acoustic or visual modalities can lead to inaccurate detection of device-directedness, resulting in spurious commands or misinterpreted utterances. This work tackles these limitations by introducing multiple modalities in a pre-trained text-only LLM \textit{efficiently}, leading to a multimodal LLM that uses additional contextual signals from audio and video data in addition to text.

One natural extension towards multimodal LLMs from text-based LLMs that several recent works have explored is to start with a pre-trained text-only model \cite{devlin-etal-2019-bert,radford2019language,NEURIPS2020_1457c0d6,raffel2020exploring,lewis-etal-2020-bart} and continue pre-training on pairs of modalities like image-text \cite{tan-bansal-2019-lxmert,lu2019vilbert,li2019visualbert,cho2021unifying}, video-text \cite{sun2019videobert,zhou2020unified,lin2023video} or audio-text \cite{elizalde2023clap,deshmukh2024pengi,gong2023listen,wagner2023multimodal} leading to multimodal image/video/audio-text LLMs. There are two limitations of this approach: (1) multimodal pre-training is compute intensive. For example, it takes 2,000 and 5,000 V100 GPU hours to pre-train on audio \cite{bertasius2021space} and video \cite{huang2022masked} data. (2) For effective pre-training, large amounts of paired data across all modalities is required, which is hard to mine and expensive to annotate at scale. Indeed, there are relatively fewer works that attempted joint audio-video-text multimodal pre-training \cite{zhang2023video,lyu2023macaw,wu2023next}. More recent works train with interleaved sequences of image/text data \cite{alayrac2022flamingo,driess2023palme} not requiring explicitly paired modalities. However, for DDSD task, aligned modalities introduce important correlations that better inform user's intent. Further, despite effective pre-training, fine-tuning would still be necessary to boost model performance \cite{NEURIPS2020_1457c0d6} for a user facing task like DDSD. 

\begin{figure}
    \centering
    \includegraphics[width=0.49\textwidth]{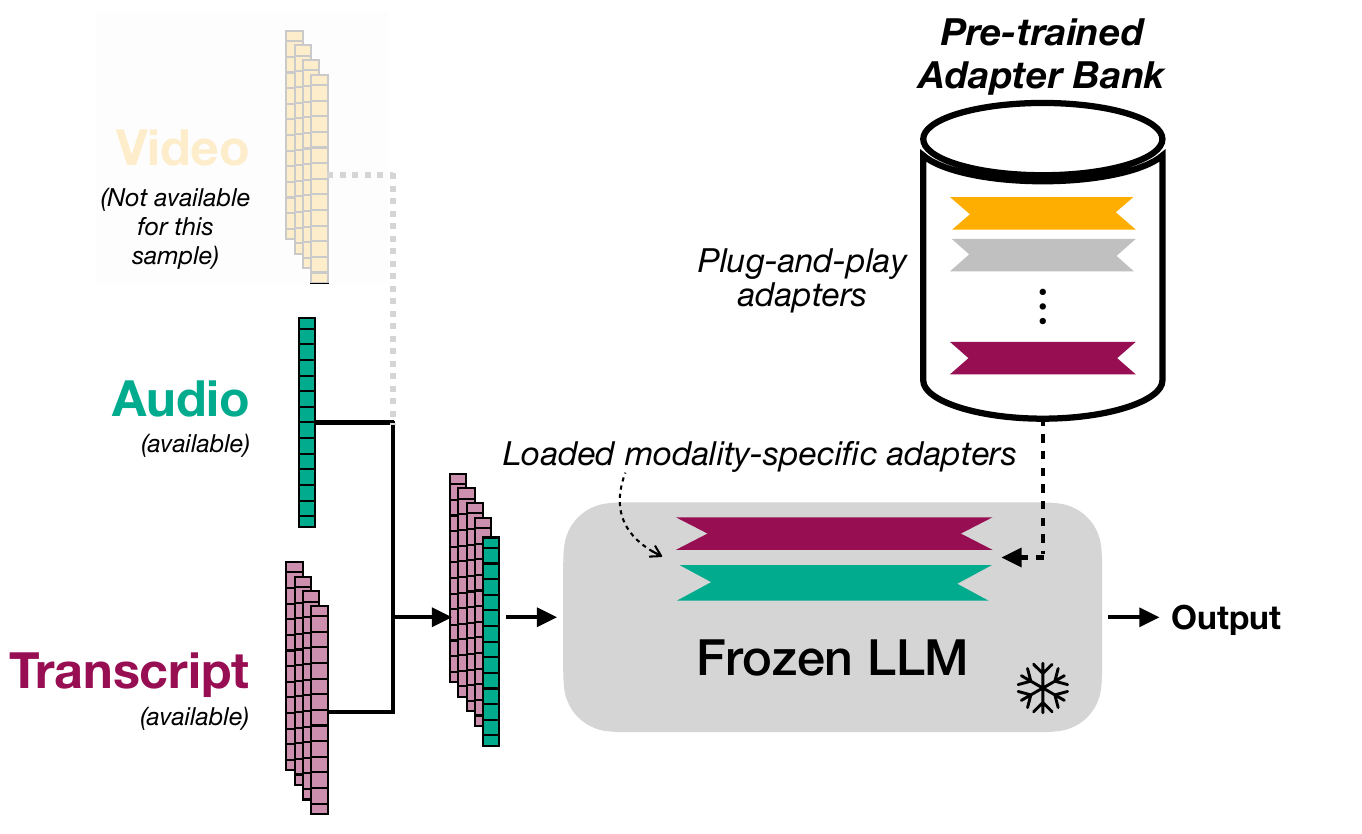}
    \caption{High-level architecture showing proposed FLoRA adapters. A frozen LLM loads modality-specific adapters to take advantage of all available modalities at training or inference time. The architecture is robust to missing modalities by dropping corresponding adapters (video in the diagram above).}
    \label{fig:illustration}
\end{figure}

\begin{figure*}
    \centering
    \includegraphics[width=\textwidth]{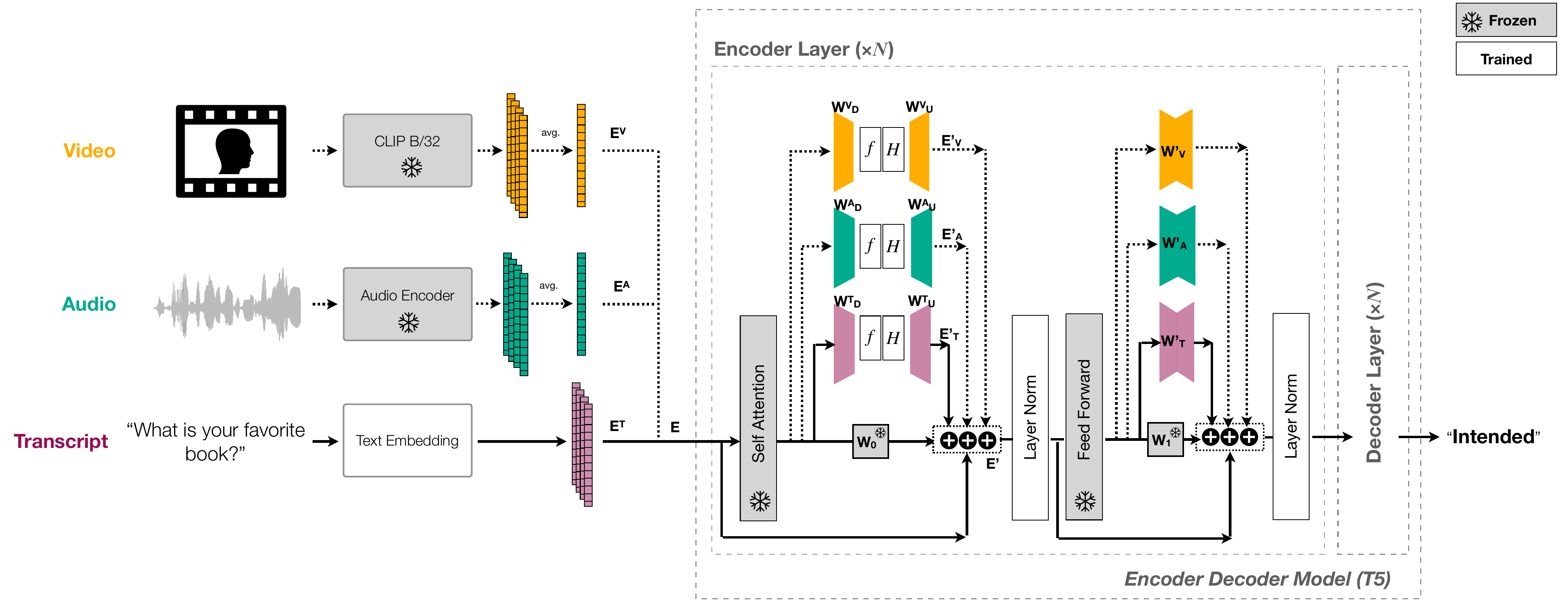}
    \caption{Model architecture showing modality-specific FLoRA layers in an encoder-decoder model. Dotted lines across audio and video modalities are optional and the corresponding adapters are trained (during training) or invoked (during inference) only if the modality is available.}
    \label{fig:technical_model_architecture}
\end{figure*}

To circumvent above limitations, recent approaches have proposed to adapt text-only LLMs by updating a small fraction of total model parameters. Most approaches focused on training smaller front-end/prefix/prompt-tuned networks \cite{lester-etal-2021-power,li-liang-2021-prefix} that attempt to learn a mapping from a new modality to the text embedding space of the LLM \cite{sun2019videobert,zhou2020unified,li2019visualbert,cho2021unifying,lin2023video}. This is a multimodal equivalent of prefix tuning approaches that were proposed to adapt text-only LLMs \cite{li-liang-2021-prefix}. Separately, for text-only LLMs, Low Rank Adaptation (LoRA) \cite{HuEJ2022lora} was introduced to efficiently adapt a pre-trained LLM to new text-only tasks and domains. LoRA was shown to achieve improved performance for new tasks compared to prefix tuning \cite{HuEJ2022lora}. MAGMA \cite{EichenbergC2022magma} and VL-adapter \cite{sung2022vl} explored such adaptation method to introduce image/video modality in pre-trained text-only LLMs for text generation tasks and MAGMA showed its benefits over prefix tuning alone explored in Frozen \cite{tsimpoukelli2021multimodal}. However, all of these approaches rely on 100\% availability of paired data across all modalities (image/video+text in their case) during both training and inference and include only one additional modality. 

Building on this line of work, in this paper, we propose \textit{Fusion} Low Rank Adaptation (FLoRA), an extension of LoRA, where we use adapters to fuse new, unseen modalities, into a pre-trained LLM, introducing a way to extend a text-based LLM to its multimodal counterpart (refer to Figure \ref{fig:illustration}). Similar to LoRA, FLoRA also requires tuning a fraction of the total model parameters enabling multimodal adaptation with limited amount of new data compared to full fine-tuning. We show the benefits of FLoRA in introducing more than one new modality. Moreover, with a frozen text-only LLM representing comprehensive world knowledge, the advantages of the proposed approach extend to unifying various tasks a VA handles, through training multiple adapters linked to a single backbone, instead of relying on numerous specialized models.

We additionally introduce \textit{adapter dropout} where modality-specific adapters can be dropped when the corresponding modality is absent for given input. This flexibility eliminates the need for massive amounts of paired data containing all modalities. We show that using FLoRA for DDSD, the resulting multimodal LLM achieves 22\% lower equal error rate (relatively) over the unimodal (text-only) approach, while reaching parity with its full fine-tuning counterpart by training only 1-5\% total model parameters. Furthermore, with adapter dropout, FLoRA is robust to missing data. When missing video data, FLoRA improves over FFT by 20\% relative reduction in equal error rate and 56\% relative reduction in false accept rate at 10\% false reject rate. Finally, we show the proposed technique scales well for model sizes from 16M to 3B parameters showing promise for resource-constrained on-device VA applications as well as for larger multimodal models.
\section{Approach}
\label{sec:approach}

Low Rank Adaptation (LoRA) \cite{HuEJ2022lora} is a well established technique as an efficient way to fine-tune LLMs on a specific downstream task. Adapter \cite{houlsby2019parameter} represents a small set of parameters added to layers of a pre-trained model, which allows to achieve as high accuracy as full fine-tuning, by fine-tuning only a small set of parameters. LoRA proposes to train adapter modules ($W'$) on top of a pre-trained LLM ($W$) by updating only the adapter parameters. During inference, $\Delta$$W$ = $W$ + $W'$ represents the updated LLM weights. The parameter complexity of $W'$ is much lower than W by enforcing a low rank structure. Typically, the chosen rank of $W'$ leads to additional adapter parameters being less than 5\% of the original LLM parameters.

\subsection{Fusion LoRA} With Fusion Low Rank Adaptation, we extend this idea to include previously unseen multimodal inputs via adapter modules, where $W'$ now learns a domain shift from text-only data to aligned, multimodal signals. Similar to prior work, we train small prefix networks that map video ($E^V$) and audio ($E^A$) representations to the text embedding ($E^T$) dimension which are then concatenated along the temporal text dimension. We use a BART \cite{lewis-etal-2020-bart} and T5 \cite{raffel2020exploring} based encoder-decoder architectures as the pre-trained language-based backbone model. Within each encoder and decoder layer, adapter layers are added after the self-attention layer and the feed-forward layer; adding two adapter layers per encoder/decoder layer. Figure \ref{fig:technical_model_architecture} shows details of this architecture.

\subsection{Modality-specific Adapters}
FLoRA can be extended to multiple adapters, one adapter per each unseen modality, task or dataset. There are no theoretical constraints on the number of adapters that could be trained to use the original LLM weights $W$ ($W_0$, $W_1$, $\dots$). As each additional adapter will increase the number of tunable parameters, it is preferable to keep this number low for efficiency. A modality specific adapter follows the FLoRA technique, but with an individual adapter for each modality. Considering a model being trained on audio (A), video (V), and text (T) modalities, updated LLM weights would include weights of each separate modality:

\begin{equation}
    E_{V}' = f(E \times W_D^V) \times H \times W_U^V
\end{equation}

\begin{equation}
    E' = (E \times W) + E'_V + E'_A + E'_T  
\end{equation}
    
where $E$ represents the multimodal input embedding, $E_V'$ represents the video-adapter representation with adapter activation function $f$, hidden layer $H$, down-sample matrix weights $W_D^V$ and up-sample matrix weights $W_U^V$. Updated embedding with modality-specific adapters $E'$ is a combination of frozen LLM representation ($E \times W$) with the video ($E_{V}'$), audio ($E_{A}'$) and text adapter ($E_{T}'$) representations.

\subsection{Adapter Dropout}
With multimodal data, it is challenging to find paired data across all samples. With the increase in number of modalities, this challenge only intensifies in nature. Multimodal LLMs require access to large amounts of data, not just for pre-training but also for fine-tuning an effective model. FLoRA addresses the latter problem by proposing to fine-tune only a small number of parameters, however, the original issue persists. For this, we propose adapter dropout. The idea is simple: if a modality, say audio, exists for a given sample point, we train the ``audio-adapter'' on that data point. If it is absent, we skip the parameter updates corresponding to that modality (audio-adapter), and train only the other modalities available (video- and text-adapter in this example). At inference time, only the adapters corresponding to modalities present in the test datasets are loaded onto the model for inference. The original pre-trained model stays unaffected throughout. FLoRA enables us to utilize any and all modalities we have available for a single data point without removing any data. At inference time, it also opens up an opportunity to test on datasets that might not have all these modalities, encouraging generalizable tests on more tasks and datasets.

\section{Experimental Setup}
\label{sec:experimental_setup}

\subsection{Classification via Generation} We evaluate our approach on the DDSD task which is inherently a binary classification problem. To make the best use of LLMs, we pose classification as a language generation problem (intended=Yes, unintended=No) to enable other generation tasks such as ASR or translation jointly. We do not employ any specific techniques to restrict model output to only these two tokens and observe a well-trained model learns to only generate these two tokens. For DDSD, we report two classification metrics: Equal Error Rate (EER) and False Accept Rate at 10\% False Reject Rate (FA@10), as they are more relevant for this task. Tokenizers for each model size are the default HuggingFace \cite{wolf2019huggingface} tokenizers with corresponding vocabulary.

\subsection{Datasets} We train and evaluate on datasets sourced for the task of DDSD using voice, touch, and back-to-back (B2B) invocation in single-turn and multi-turn dialog. Voice-based invocation involves the trigger keyword \textit{Siri} or \textit{Hey Siri}, touch-based invocation is keyword-less, and B2B is a mix of both. We use three datasets with varying modalities, invocation style, and difficulty levels. \datasethobbit\ is an in-house multimodal (audio, video, and text) dataset of users engaging with both a VA and other users in a multi-turn dialog. \datasetkorra\ is also curated for user-device interaction. This dataset contains only audio recordings and is used only for testing in this work. Finally, \datasetthk, is also an audio-based dataset of in-the-wild usage of B2B. This is not a curated dataset and involves long contextual dialog which is a mixture of human-human and human-device speech, making it the most challenging set. All datasets are in English containing diverse demographics of opt-in users and natural conversation. \datasethobbit\ has a confidence interval of +/- 0.3\% and \datasetkorra\ and \datasetthk\ of +/- 1\%. Table \ref{tab:dataset_statistics} summarizes dataset details.

\begin{table}[]
    \centering
    \begin{tabular}{ccccc}
    \toprule
        \textbf{Dataset} & \textbf{Modalities} & \textbf{\# Train} & \textbf{\# Test} & \textbf{Characteristic} \\
        \midrule
         \datasethobbit & A,V,T & 120K & 40K & Curated \\
         \datasetkorra  & A,T   & - & 7K & Curated \\
         \datasetthk    & A,T   & 1M & 14K & in-the-wild \\
    \bottomrule
    \end{tabular}
    \caption{Dataset overview and statistics for B2B datasets. Audio (A), video (V) and text (T) modalities vary across the three. }
    \label{tab:dataset_statistics}
\end{table}

\subsection{Baselines and Toplines} The audio features are extracted via a pre-trained acoustic false trigger mitigation model \cite{rudovic2022streaming} trained jointly for speech recognition and DDSD. We extract a single 256 dimension vector per utterance by mean pooling across the frame-wise embeddings from the last layer of the encoder. The video features are extracted via a pre-trained CLIP model \cite{radford2021learning}, ViT-B/32, by sampling videos at 10 frames per second and averaging features across an utterance to get a single 512 dimension vector. As baselines, we train unimodal models on top of the features extracted from pre-trained models by training a fully-connected linear layer mapping modality-specific embeddings to binary labels for each dataset. For the text baseline, we fine-tune the T5/BART language model only with text inputs. For topline, we fine-tune a well established multimodal LLM VL-BART \cite{cho2021unifying} for the DDSD task. VL-BART is originally proposed as a image/video+text model. We extend their approach to additionally include audio. Fully fine-tuned (FFT) models represent an ideal upper bound, i.e., best performance possible from large models if all modalities were available with the full parameter set. 

All LLMs are initialized by their corresponding weights downloaded from HuggingFace. Across models, a rank of 4 is used in the adapters. We update the layernorm parameters, with a learning rate of 1e-5 with adamw optimizer \cite{loshchilov2017decoupled}, with warm up ratio 0.1, batch size 256 for most model sizes with upto 4 epochs of training on a single 32GB GPU. For T53B models, we use deepspeed \cite{rasley2020deepspeed} to optimize model training within 32GB GPU memory across 8 GPUs each. 
\section{Results and Discussion}
\label{sec:results}

\begin{table*}[t]
\centering
\begin{tabular}{llccccccccc}
\toprule
\multicolumn{2}{c}{} & \textbf{Trainable} & & \multicolumn{2}{c}{\textbf{\datasethobbit}} & \multicolumn{2}{c}{\textbf{\datasetkorra}} & \multicolumn{2}{c}{\textbf{\datasetthk}} \\
\textbf{Category} & \textbf{Method} & \textbf{Params} & \textbf{Signal} & \textbf{EER} & \textbf{FA@10} & \textbf{EER} & \textbf{FA@10} & \textbf{EER} & \textbf{FA@10} \\ 
\midrule
\multirow{2}{*}{Unimodal} & FFT  & 100\%         & T    & 8.8  & 7.0  & 12.8 & 15.8 & 14.0 & 20.2 \\
\multirow{2}{*}{Baselines} & FFT  & 100\%         & A    & 11.0 & 13.1 & 11.6 & 14.0 & 10.3 & 10.9 \\
& FFT  & 100\%         & V    & 33.9 & 64.0 & -   & -   & -    & - \\
\midrule
\multirow{1}{*}{VL-BART (fine-tuned)} & FFT  & 100\% & A+V+T  &  \textbf{6.5} & \textbf{1.9}  & 15.4  & 25.0 & 14.3 & 22.7 \\
\midrule
\multirow{1}{*}{Proposed Approach} & FLoRA & 5\%   & A+V+T& 6.8 & 2.6 & \textbf{10.8} & \textbf{12.2} & \textbf{8.4} & \textbf{6.9} \\
\bottomrule
\end{tabular}
\caption{Comparing FLoRA models for new modalities; Audio (A), Video (V), Text (T) with full fine-tuned (FFT) unimodal models and corresponding multimodal LLM models.} 
\label{tab:main_results}
\end{table*}

\subsection{Fusion via Low Rank Adaptation}

Table \ref{tab:main_results} presents the main results of the proposed approach. We compare FLoRA against full fine-tuning of a multimodal LLM VL-BART \cite{cho2021unifying}. We fine-tune VL-BART \cite{cho2021unifying} on internal data for DDSD and additionally add the audio modality in the same spirit. This model acts as an effective topline for this work demonstrating best possible performance if all parallel data were available. FLoRA models perform significantly better than unimodal baselines across all datasets, for instance on \datasethobbit\ 22\% improvement in EER and 62\% improvement in FA@10 over the best unimodal baseline (text-only LLM). As video-based data is available only for \datasethobbit, FFT performance is most notable on that, whereas for \datasetkorra\ and \datasetthk, performance drops due to the missing video signal. Correspondingly, FLoRA models are close in performance to FFT on \datasethobbit\ and perform much better on the other two datasets achieving 50-70\% relative improvement in FA@10, demonstrating their robustness to missing video modality. Further, FLoRA models train only 5\% of total model parameters. These results demonstrate we can inject an unseen modality into a pre-trained LLM just by training modality-specific adapters with efficiency and robustness. 

\subsection{Adapter Dropout}
To demonstrate adapter dropout, we simulate a missing modality setup comparing effects of missing data on FFT and FLoRA. For the \datasethobbit\ test set, we remove audio and video data one modality at a time. Table \ref{tbl:adapter_dropout} summarizes the results. The FFT approach, while powerful when all data is present, cannot adapt to missing data. FLoRA outperforms FFT by 14-20\% relative improvement in EER and 38-56\% relative improvement in FA@10 (refer Table \ref{tbl:adapter_dropout}). With missing video or audio, FFT model performance results in 38\% relative degradation in EER over FFT having access to all data (EER 9.0 vs. 6.5, refer Tables \ref{tbl:adapter_dropout} and \ref{tab:main_results}). Correspondingly, the FLoRA model is more robust to missing modalities with a relative degradation of 6\% for missing video and 13\% for missing audio, see Table \ref{tbl:adapter_dropout}.

\begin{table}[t]
\centering
\begin{tabular}{ccccc}
\toprule
\multicolumn{1}{c}{} & \multicolumn{2}{c}{\textbf{Missing Video}} & \multicolumn{2}{c}{\textbf{Missing Audio}} \\
\textbf{Modalities}  & \textbf{EER}  & \textbf{FA@10} & \textbf{EER}  & \textbf{FA@10} \\ 
\midrule
FFT A+V+T  & 9.0 & 7.3 & 9.0 & 6.9 \\
FLoRA A+V+T    & \textbf{7.2} & \textbf{3.2} & \textbf{7.7} & \textbf{4.3} \\
\bottomrule
\end{tabular}
\caption{Effect of missing modalities on FFT and FLoRA approaches showing the versatility of adapter dropout on \datasethobbit\ test set.} 
\label{tbl:adapter_dropout}
\end{table}

\subsection{Scaling Model Size}
\begin{figure}
    \centering
    \includegraphics[width=0.43\textwidth]{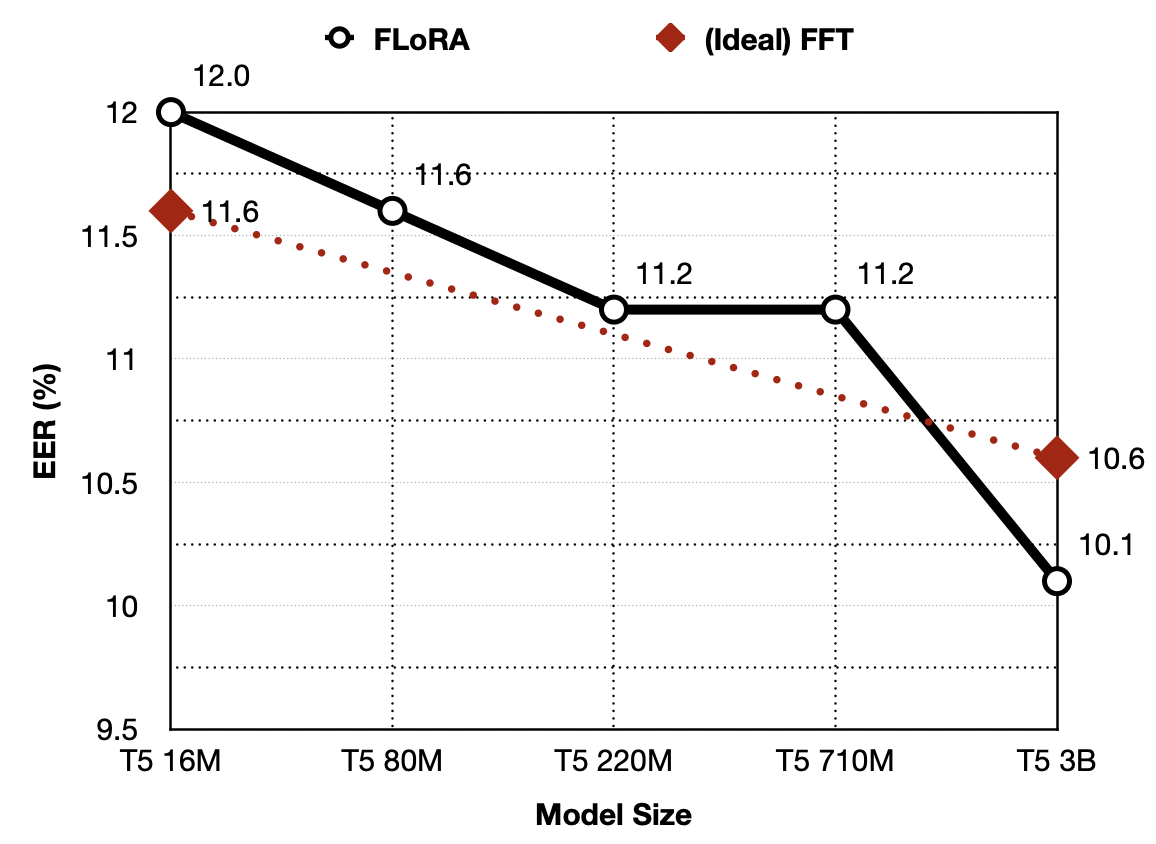}
    \caption{Model scalability for model sizes from 16M parameters to 3B parameters for the T5 architecture. The FLoRA technique scales across all model sizes with performance on \datasetthk\ improving consistently with increase in model size as expected.}
    \label{fig:scalability}
\end{figure}

Finally, we study the effect of increasing and decreasing LLM model size on the DDSD task. We report results on \datasetthk, which is the largest and most difficult test set. Figure \ref{fig:scalability} shows a consistent improvement in EER with the increase in model size from 16M parameters to 3B parameters as expected. All models are T5 variants. Depending on model size the FLoRA parameters are within 0.8\% to 5\% of total parameters. For T516M, FFT performance is better but as we increase model size, T53B FLoRA models perform better. This could be attributed to FFT models overfitting on this task as the model size increases, forgetting their generalizable pre-training when updating all 3B parameters for the DDSD task. In constrast, FLoRA models only update 64M parameters for the T53B model, and are hence able to retain generalizability. Model parameter scaling results show a promising trend for large-scale multimodal LLMs or small scale on-device friendly models for limited paired data.

\section{Conclusion}
\label{sec:conclusion}

We introduced FLoRA, a method to adapt a text-only LLM to previously unseen modalities by training low rank adapter modules. We demonstrate that this approach comes close in performance or improves over the corresponding full fine-tuned multimodal LLM for DDSD. The proposed approach is parameter efficient updating at most 1-5\% of total model parameters and scales from small LMs of 16M parameters to large LMs of 3B parameters. Moreover, an important aspect of the proposed approach is new adapter dropout technique, a method to plug-and-play pre-trained adapters to efficiently handle missing modalities during train or test time. When tested on missing modality setup, FLoRA models outperform full fine-tuned models by 20\% relative improvement in EER.
\section{Acknowledgements}
We thank Pranay Dighe, Vineet Garg and Zakaria Aldeneh for feedback on the paper.

\bibliographystyle{IEEEtran}
\bibliography{mybib}

\end{document}